\documentclass[sigconf]{acmart}

\usepackage{booktabs}
\usepackage{array}
\usepackage{amsmath}
\usepackage{url}
\usepackage{graphicx}

\copyrightyear{2026}
\acmYear{2026}
\setcopyright{cc}
\setcctype{by}
\acmConference[ICAIF '26]{7th ACM International Conference on AI in Finance}{November 14--17, 2026}{Milan, AA, Italy}
\acmBooktitle{7th ACM International Conference on AI in Finance (ICAIF '26), November 14--17, 2026, Milan, AA, Italy}
\acmDOI{10.1145/3858063.3858690}
\acmISBN{979-8-4007-3087-0/2026/11}

\begin{document}

\title{Sparse Autoencoders as Implicit Diversification Regularisers for
Cross-Sectional Cryptocurrency Returns}

\author{Weijia Han}
\orcid{0009-0000-8300-6691}
\affiliation{
  \institution{University of Washington}
  \city{Seattle}
  \state{Washington}
  \country{USA}}
\email{weivivi77@gmail.com}

\author{Lisha Qu}
\orcid{0009-0001-5393-6991}
\affiliation{
  \institution{University of Washington}
  \city{Bothell}
  \state{Washington}
  \country{USA}}
\email{lishaq@uw.edu}

\begin{abstract}
Factor models are central to asset pricing and portfolio construction, but
neural factors are often difficult to interpret, and their effect on portfolio
risk is unclear. A sparse autoencoder (SAE) may address both concerns by
describing daily returns across many assets with only a few active features.
We ask whether this sparse representation produces stable financial factors
and better-diversified portfolios.
We study cryptocurrency returns, use equities as a secondary test, and compare
the SAE with common benchmarks in a time-ordered evaluation. We find
consistently lower volatility in SAE portfolios and link this result to lower
correlations among the selected assets. However, the SAE does not outperform a
simple contrarian strategy on risk-adjusted return, and its factor identities
are not stable across training runs.
Our contribution is to show that sparse coding can support portfolio
diversification, while clearly separating this risk effect from claims about
alpha or stable factor identities.
\end{abstract}



\maketitle

\section{Introduction}
\label{sec:intro}

Factor models are the organising framework of empirical asset pricing.
From the Capital Asset Pricing Model (CAPM)~\cite{sharpe1964capital} to the Fama-French five-factor
model~\cite{fama1993common,fama2015five}, the premise is that cross-sectional
return variation decomposes into a handful of systematic risk exposures plus
idiosyncratic noise.
The rise of machine learning has relaxed the assumption of linearity:
\citet{gu2020empirical} show that feed-forward networks and gradient-boosted
trees consistently outperform linear factor models in equity return prediction,
and \citet{kelly2019characteristics} demonstrate that an instrumented form of
principal component analysis (PCA) bridges
characteristic-based and latent factor representations.
Yet predictive gains carry an interpretability cost: a neural network encodes its
implicit ``factors'' as distributed activation patterns across thousands of
weights, frustrating any clean mapping between network dimensions and
identifiable economic forces.
The obstacle is structural: \citet{elhage2022superposition} show that neural
networks systematically encode more features than they have dimensions,
representing distinct concepts in linear superposition within each neuron.
Applied to financial returns, superposition implies that each latent dimension
of a dense autoencoder conflates multiple economic forces.

\paragraph{Sparse autoencoders (SAEs) as a candidate.}
\citet{bricken2023monosemanticity} and \citet{huben2024sparse} show
that an $\ell_1$ penalty on the latent layer of an autoencoder can recover
\emph{monosemantic} features: latent dimensions that activate for one coherent
concept rather than a mixture of many. \citet{templeton2024scaling}
show that this approach scales to large language models.
We test whether the same principle transfers from token embeddings to asset
returns.
If so, monosemanticity-style SAEs would provide a factor layer whose
dimensions are individually attributable to economic themes.
We find no evidence of seed-reproducible factor identities
(\S\ref{subsec:interp}); what the bottleneck does deliver is a change in the
risk of the portfolios built on it, and that is the subject of this paper.

\paragraph{Cryptocurrency markets as a testbed.}
Crypto markets trade 24/7, include more than $400$ assets with high cross-sectional
return dispersion, and have fewer documented factors than
equities; \citet{liu2022crypto} document momentum, size, and network premia.
Because that dispersion inflates long-short Sharpe magnitudes, we treat
absolute crypto Sharpes as within-panel, cross-method comparisons and perform
a secondary transfer check on $200$ S\&P~500 stocks.

\begin{figure*}[!t]
  \centering
  \includegraphics[width=\linewidth]{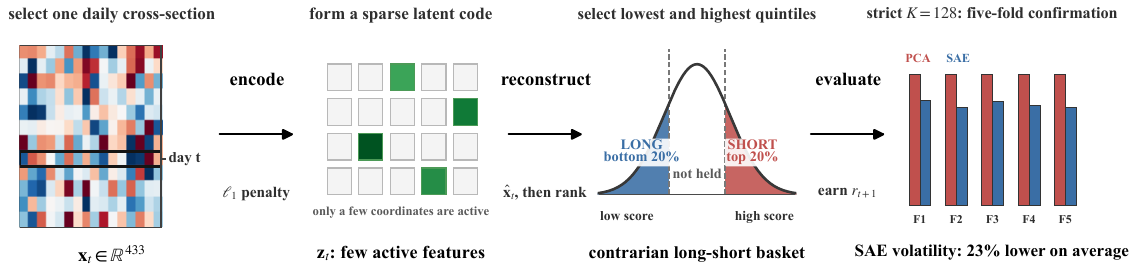}
  \caption{Pipeline from the daily crypto return cross-section
    $\mathbf{x}_t$ through a sparse code and reconstructed-return ranking to a
    contrarian quintile basket. Under strictly pre-fold selection, the chosen
    $K{=}128$ SAE reduces rank-scale long-short volatility by $23\%$ on average
    relative to matched-$K$ PCA ($20$--$26\%$ in all five folds).}
  \label{fig:overview}
  \Description{Pipeline from a daily cross-section of 433 cryptocurrency
    returns through a sparse code and a contrarian long-short basket. The final
    bars show lower confirmed rank-scale volatility for SAE than PCA.}
\end{figure*}

\paragraph{Contributions.}
We study a portfolio-construction property: the realised risk of baskets
ranked on a factor model's reconstruction.
We make no asset-pricing, forecasting, or alpha claim.
At the exploratory $K{=}64$ setting, individual factor identities are
not seed-reproducible: retraining with
another seed yields an essentially different matched dictionary
(\S\ref{subsec:interp}).
Figure~\ref{fig:overview} summarises the end-to-end pipeline.
\textbf{(C1)}~We provide a walk-forward empirical study of a
monosemanticity-style $\ell_1$ autoencoder for cross-sectional factor
discovery. The model is a standard sparse-coding or dictionary-learning
estimator~\cite{olshausen1996emergence,mairal2009online}. Three protocol changes
reduce common sources of bias in out-of-sample evaluation: a
validity-masked reconstruction loss (which avoids bias toward observation
patterns from zero-filling missing returns), chronological splits, and a
cross-sectional clipped inverse-normal rank transformation
(\S\ref{subsec:data}).
\textbf{(C2)}~Under strictly pre-fold selection, the chosen $K{=}128$ SAE has
$20$--$26\%$ lower long-short volatility than matched-$K$ PCA in all five
confirmation folds. Exploratory $K{=}64$ analyses show the same direction on
raw returns and associate it with lower within-basket pairwise
correlation ($0.008$ vs $0.039$, $p{<}0.001$), not selection of
low-volatility assets.
A Ledoit--Wolf minimum-variance overlay on the naive contrarian reduces
the SAE-versus-naive volatility gap by approximately $73\%$. The SAE remains
less volatile in four of five folds, but the remaining difference is not
statistically significant at the $5\%$ level under a two-sided test
($p{=}0.057$).
\textbf{(C3)}~The volatility reduction also appears in a secondary exploratory
equity analysis, where the lowest realised volatility is $6.6\%$. The equity
factors have low pairwise correlations with the Fama--French five-factor model
plus momentum (FF5+MOM; max $|\rho|{=}0.147$). However, the cryptocurrency
Sharpe ranking does not transfer: the equity SAE--PCA walk-forward Sharpe gap
is not statistically significant ($-0.20$, $p{=}0.160$), and the SAE also
trails the naive contrarian ($-0.66$, $p{=}0.072$).
The naive contrarian achieves the highest pre-cost crypto Sharpe on the rank scale
(gap $-2.91$, $p{=}0.002$) and in money units ($-2.48$, $p{=}0.009$), and a
lagged-return multilayer perceptron (MLP) in the spirit of
\citet{gu2020empirical} widens the gap.
We position the SAE as a bottleneck-induced diversification regulariser.

\section{Related Work}
\label{sec:related}

Factor models for cross-sectional returns range from the Fama-French
families~\cite{fama1993common,fama2015five} and data-driven linear
alternatives (PCA, instrumented PCA~\cite{kelly2019characteristics}, and
risk-premium PCA~\cite{lettau2020factors}) to nonlinear machine-learning methods that
improve return-prediction accuracy at the cost of interpretability:
\citet{gu2020empirical} show that gradient-boosted trees and feed-forward
networks outperform factor models in large equity panels, while
\citet{chen2024deep} learn deep conditional factors from firm characteristics;
the conditional autoencoder of \citet{gu2021autoencoder} is the closest
architectural ancestor, but its latent loadings are characteristic-driven,
without an $\ell_1$ penalty.
In natural language processing (NLP), SAEs resolve
\emph{superposition}~\cite{elhage2022superposition} by enforcing $\ell_1$
sparsity on transformer activations; \citet{bricken2023monosemanticity} and
\citet{huben2024sparse} demonstrate that the resulting features are
monosemantic. \citet{templeton2024scaling} extend this approach to large
language models.
In finance, \citet{kozak2020shrinking} apply $\ell_1$ shrinkage to stochastic
discount factor (SDF) pricing
coefficients (sparsity on weights, distinct from codes), and \citet{avellaneda2010stat}
establish the residual-reversal statistical-arbitrage pipeline that motivates
our reconstruction-based long-short ranking.
For the crypto testbed, \citet{liu2022crypto} document momentum, size, and
network premia that motivate future crypto-native spanning tests.
To our knowledge, the NLP monosemanticity framework has not been applied to
cross-sectional factor discovery.

\section{Method}
\label{sec:method}

\subsection{Architecture and loss}
\label{subsec:arch}

Let $\mathbf{x}_t \in \mathbb{R}^N$ denote the cross-section of preprocessed
returns on day~$t$, with $N{=}433$ (crypto) or $N{=}200$ (equity).
The SAE maps each cross-section through a rectified linear unit (ReLU) encoder
to a $K$-dimensional nonnegative sparse latent code and reconstructs it through
a linear decoder:
\begin{equation}
  \mathbf{z}_t = \operatorname{ReLU}\!\left(
    \mathbf{W}_{\mathrm{enc}}\,\mathbf{x}_t + \mathbf{b}_{\mathrm{enc}}
  \right),\quad
  \hat{\mathbf{x}}_t = \mathbf{W}_{\mathrm{dec}}\,\mathbf{z}_t + \mathbf{b}_{\mathrm{dec}},
  \label{eq:sae}
\end{equation}
where $\mathbf{W}_{\mathrm{enc}} \in \mathbb{R}^{K \times N}$,
$\mathbf{W}_{\mathrm{dec}} \in \mathbb{R}^{N \times K}$,
$\mathbf{b}_{\mathrm{enc}} \in \mathbb{R}^{K}$, and
$\mathbf{b}_{\mathrm{dec}} \in \mathbb{R}^{N}$.
Decoder columns are unit-$\ell_2$-normalised so that
$\|\mathbf{z}_t\|_1$ is a meaningful proxy for total factor usage; all
reported runs tie the decoder to the encoder
($\mathbf{W}_{\mathrm{dec}}{=}\mathbf{W}_{\mathrm{enc}}^{\top}$).
The training objective combines cell-pooled validity-masked
reconstruction error with a per-day mean $\ell_1$ penalty over training
days $\mathcal{T}_{\mathrm{tr}}$:
\begin{equation}
  \mathcal{L} =
  \frac{\sum_{t \in \mathcal{T}_{\mathrm{tr}}} \sum_{i \in \mathcal{M}_t}
    (x_{ti} - \hat{x}_{ti})^2}{\sum_{t \in \mathcal{T}_{\mathrm{tr}}} |\mathcal{M}_t|}
  + \frac{\lambda}{K} \cdot
  \frac{1}{|\mathcal{T}_{\mathrm{tr}}|}\sum_{t \in \mathcal{T}_{\mathrm{tr}}}
    \|\mathbf{z}_t\|_1,
  \label{eq:loss}
\end{equation}
where $\mathcal{M}_t$ is the set of assets with observed returns on
day~$t$ and $\lambda$ is the sparsity coefficient.
Reconstruction error is pooled over \emph{observed cells}, so days with more
valid assets receive proportionally more weight.
The validity mask is central to our protocol. Without it, zero-filled returns
could make the decoder learn missingness patterns rather than return structure.
We set missing encoder inputs to zero so that the linear map is defined, but
exclude those entries from the loss. At test time, we also treat residual
missing returns as zero (flat) in the rank-scale portfolio evaluation.
Because a missing return may reflect an asset collapse, and the contrarian
strategy buys recent losers, \S\ref{subsec:portfolio} tests this convention
with a tradability mask and a separate loss calculation for assets that
disappear from the data.
Training uses Adam ($K{=}64$, $\lambda{=}10^{-2}$, $100$ epochs on the
static split), retaining the checkpoint with minimum validation loss;
validation loss is still declining at the epoch cap, so the retained
checkpoint coincides with the final epoch.

\subsection{Data and preprocessing}
\label{subsec:data}

Crypto data are Binance daily open, high, low, close, and volume (OHLCV)
observations from
2019-01-01 to 2024-12-31, covering
$N{=}433$ assets over $1{,}988$ days after a coverage filter (assets must have
$\ge 252$ non-missing days; days with fewer than $50$ valid prices are
dropped). The crypto pairs are quoted against Tether (USDT) and use
Coordinated Universal Time (UTC) day boundaries;
stablecoins, leveraged tokens, and duplicate asset representations are
excluded.
Equity data are S\&P~500 daily adjusted-close returns (Yahoo Finance,
2015-01-01 to 2024-12-31), starting from a static list of large constituents
compiled in early 2025 and retaining the $N{=}200$ stocks that pass the same
minimum-observation filter. This is not a point-in-time constituent universe,
so the equity analysis carries survivorship exposure; we accordingly treat it
as a secondary exploratory transfer check. Both panels end on 31 December
2024, and all reported experiments use this common sample period.
The winsorised ($1$st$/99$th percentiles per day) close-to-close log-returns
$\mathbf{r}_t$ are mapped cross-sectionally to clipped inverse-normal rank
scores. We use average ranks over the observed set $\mathcal{M}_t$, following
the van der Waerden approach~\cite{vanderwaerden1952rankit}. This differs from
the classical rankit $(\operatorname{rank}{-}0.5)/N$. The resulting vector is
the SAE input $\mathbf{x}_t$ in Eq.~\ref{eq:sae}:
\begin{equation}
  x_{ti} = \Phi^{-1}\!\left(
    \operatorname{clip}\!\left[
      \frac{\operatorname{rank}_i(\mathbf{r}_t)}{N_t},\, 0.01,\, 0.99
    \right]
  \right),
  \label{eq:rankit}
\end{equation}
where $N_t{=}|\mathcal{M}_t|$ is the number of observed assets on day~$t$
and the clip bounds the tails of the inverse normal.
Data are split \emph{chronologically} ($70/10/20$ train/val/test, no
shuffling).
Residual missing values are forward-filled with a strict bound of three
consecutive days; longer gaps remain missing, are masked in the SAE
loss (Eq.~\ref{eq:loss}), and are zero-filled at evaluation as
disclosed in \S\ref{subsec:arch}.

\paragraph{Survivorship and point-in-time limitations.}
The coverage filter uses full-sample information and is therefore not fully
point-in-time (PIT). Delisted assets remain in the panel through their final
quote and are masked afterwards.
As a survivorship check, we repeat the headline SAE--PCA comparison with a
fold-specific eligibility rule inside the ex-post $433$-asset universe. This
does not reconstruct a true point-in-time universe, but it uses only
information available before each test fold: an asset is tradable only if at
least $126$ of the $252$ days preceding the fold's test start are non-missing
($322$--$367$ of $433$ assets qualify per fold).
The Sharpe gap remains positive (mean $+1.15 \pm 0.80$, $9/10$ folds,
paired $t(9){=}4.57$, $p{=}0.001$) and the volatility
reduction holds in $10/10$ folds (mean $34\%$); a stricter
fixed-universe PIT variant ($216$ assets) is among the subsets of
Fig.~\ref{fig:mechanism}b.

\subsection{Baselines and evaluation}
\label{subsec:baselines}

We compare three headline baselines in the body.
\textbf{PCA at matched $K$} (top-$K$ principal components via exact singular
value decomposition (SVD)) is the primary structural comparator. Matching
$K{=}64$ controls latent dimension, but PCA and the SAE still differ in
architecture, activation, objective, and missing-value treatment.
\textbf{Dense autoencoder (DenseAE)} ($\lambda{=}0$ in Eq.~\ref{eq:loss}) shares the SAE's
architecture but removes the sparsity penalty, isolating the role of $\ell_1$.
\textbf{Naive contrarian} ranks assets by same-day return and goes long the
bottom quintile and short the top quintile. Within each leg, equal weighting
follows the simple $1/N$ benchmark of \citet{demiguel2009optimal}. This is the
no-model baseline;
it has higher pre-cost rank-scale Sharpe than all factor-model methods on
this panel, previewing the no-incremental-Sharpe finding of \S\ref{subsec:scope}.
Every reconstruction model (SAE, PCA, DenseAE) applies this same quintile
rule, ranking by the model reconstruction $\hat{\mathbf{x}}_t$ (Eq.~\ref{eq:sae})
in place of the raw return: long the lowest-$\hat{x}$ quintile, short the
highest, equal-weighted.
We also include four data-driven baselines in the walk-forward comparison of
\S\ref{subsec:portfolio}: nonnegative matrix factorisation (NMF),
independent component analysis (ICA), sparse PCA, and dictionary learning.
A characteristic-based (Barra-type) factor model needs fundamentals and
market capitalisations that our crypto panel lacks, so the baselines span
only the no-model and data-driven categories.
PCA is fitted by exact SVD in every Sharpe comparison and in
Table~\ref{tab:portfolio} and Fig.~\ref{fig:pareto}; the correlation and
universe-subset analyses of Fig.~\ref{fig:mechanism} and the static equity
analysis use randomised SVD. Exact-SVD sensitivity checks change the static
basket correlation from $0.039$ to $0.040$ and the fold volatility reductions
by at most $2.3$ percentage points.

\paragraph{Evaluation protocol.}
Each observation $\mathbf{r}_t$ is a close-to-close return and is available
only after the day-$t$ close. The reported daily-close portfolio pairs the
resulting ranking with the day-$t{+}1$ close-to-close return. The evaluation
therefore assumes that positions are entered at the same close used to form
the signal. Because this trade is not executable in practice, we interpret the
results as signal evidence rather than as a tradable backtest. Trading at the
next available price could change the results, and the daily data do not
support claims about execution or capacity. The same timing convention defines the
contrarian information coefficient (IC) of \S\ref{subsec:factor-quality}.
The primary walk-forward uses $5$ expanding folds with $159$-day test
windows ($795$ evaluation days); the SAE is retrained from scratch per
fold over seeds $\{42,123,456,789,1024\}$.
A finer ten-fold expanding design ($994$-day initial training window,
$99$-day test folds, one rotated seed per fold) is used for robustness analyses
(Table~\ref{tab:robustness}).
We use paired fold-level tests together with stationary
block-bootstrap~\cite{politis1994stationary} $95\%$ intervals
($1{,}000$ resamples over paired daily return differences,
geometric block lengths with mean $\ell{=}5$ days).
The exploratory static mechanism checks instead use $10{,}000$ stationary
block-bootstrap resamples of paired daily statistic differences with mean
$30$-day blocks, matching their rolling estimation window.
With $p_t$ the daily equal-weighted long-short profit-and-loss (long and
short legs each gross $1$, zero net) and $\bar{p}, s(p)$ its sample mean
and standard deviation over test days, we report
$R_{\mathrm{ann}}{=}A\bar{p}$, $\sigma_{\mathrm{ann}}{=}\sqrt{A}\,s(p)$, and
$\mathrm{SR}{=}\sqrt{A}\,\bar{p}/s(p)$, with $A{=}365$ (crypto) or $252$
(equity) and zero risk-free rate.
The exploratory $K{=}64$, $\lambda{=}10^{-2}$ setting came from an
initial reconstruction-loss sweep whose selection window overlaps the early
walk-forward test windows (folds begin at $60\%$ of the sample). Although it
was not selected on test Sharpe, the overlap makes the $K{=}64$ portfolio
comparisons exploratory. We therefore repeat selection on the first $748$
days, strictly before all confirmation folds. The grid uses
$K{\in}\{32,64,128\}$ and
$\lambda{\in}\{10^{-3},10^{-2},10^{-1}\}$. Learning rates are
$\eta{\in}\{3{\times}10^{-4},10^{-3}\}$, giving $18$ configurations.
Validation reconstruction selects the largest model and strongest penalty:
$K{=}128$ and $\lambda{=}10^{-1}$, with $\eta{=}10^{-3}$. Loss falls
monotonically in $K$, and the original $K{=}64$ configuration ranks ninth.
When we rerun the walk-forward analysis at the selected setting, using five
$248$-day folds that begin after the selection window, the SAE reduces
volatility by $20$--$26\%$ in all five folds. The SAE--PCA Sharpe gap is
$+0.33$ ($4/5$ folds, $p{=}0.14$).
Across $K{\in}\{32,64,128,256,433\}$ under this unified protocol
($\lambda{=}10^{-1}$), the volatility reduction is positive in every fold.
For increasing $K$, the mean reductions are $32\%$, $26\%$, $23\%$, $18\%$,
and $17\%$. The corresponding Sharpe gaps are $-0.55$, $+0.41$, $+0.33$,
$-0.63$, and $-0.83$. The gap is largest at an intermediate value of $K$ and becomes
significantly negative only for $K{\geq}256$ ($p{\leq}0.039$), as matched-$K$
PCA approaches the naive contrarian when $K{\to}N$.
The volatility claim survives strict re-selection; the Sharpe ranking remains
specific to the setting.
Walk-forward retraining uses $50$ epochs (the static-split model of
\S\ref{subsec:arch} trains for $100$).
In the exploratory $K{=}64$ analysis, checkpoint selection used the same folds
that were later evaluated. Although the selected checkpoints coincided with
the final-epoch models, this design is not strictly out of sample. We therefore
treat all $K{=}64$ results as exploratory and base our main conclusion on the
strictly pre-fold $K{=}128$ analysis above.
The overlay control of \S\ref{subsec:scope} uses a trailing
$60$-day Ledoit--Wolf-shrunk covariance with $20\%$ quintiles.
All $p$-values are two-sided; where the hypothesis is directional (the
overlay residual) we add the one-sided value.

\paragraph{Interpretation of rank-scale results.}
Reported absolute Sharpes are pre-cost rank-scale artefacts of the per-day
clipped inverse-normal rank transformation (Eq.~\ref{eq:rankit}): they are valid
only for cross-method comparison. We do not subtract monetary transaction
costs from these dimensionless scores; all economic return comparisons use
the raw-return panel and are reported pre-cost.

\section{Results}
\label{sec:results}

\subsection{Factor quality at matched K}
\label{subsec:factor-quality}

At the exploratory setting $K{=}64$, the $\ell_1$ penalty substantially
increases code sparsity, while we find no clear evidence of an improvement in
predictive IC.
We measure code sparsity with the Hoyer index~\cite{hoyer2004non}:
\[
H(\mathbf{z})=
\frac{\sqrt{K}-\|\mathbf{z}\|_1/\|\mathbf{z}\|_2}{\sqrt{K}-1}.
\]
The index is evaluated on $|\mathbf{z}|$ and ranges from $0$ for an
equal-magnitude dense code to $1$ for a one-hot code. Over five seeds, it is
$0.497 \pm 0.003$ for the SAE, compared with $0.303$ for DenseAE
($\lambda{=}0$, identical architecture) and $0.272$ for PCA scores. Thus, the
dense autoencoder's latent code is not sparse on these data without the penalty.
Contrarian IC ($-\operatorname{IC}(\mathbf{r}_{t+1},\hat{\mathbf{x}}_t)$ on
the test split) is
$+0.016 \pm 0.002$ for the SAE, compared with $+0.020$ for DenseAE and
$+0.021$ for PCA. These differences are small relative to cross-day IC noise
and remain untested.
ReLU and JumpReLU~\cite{rajamanoharan2024jumping} SAE variants produce similar
values at the reported precision (IC and Hoyer match to two decimal places).

\subsection{Portfolio: volatility reduction is the robust effect}
\label{subsec:portfolio}

Table~\ref{tab:portfolio} presents exploratory contrarian quintile long-short
results at $K{=}64$ on the clipped inverse-normal rank scale of
\S\ref{subsec:baselines}. All results are pre-cost.
At the same latent dimension, the SAE realises $4.2\%$ annualised portfolio
volatility versus PCA-$K{=}64$'s $6.8\%$, a $38\%$ reduction, while
earning \emph{lower} rank-scale return ($+13.9\%$ vs $+18.1\%$).
The SAE's higher pre-cost Sharpe ($+3.28$ vs $+2.67$) therefore comes from
lower volatility, not higher return.
DenseAE ($6.4\%$ volatility) and NMF ($7.0\%$ volatility) are closer to PCA
than to the SAE. The same is true for sparse PCA and dictionary learning,
whose walk-forward mean volatilities are $7.5\%$ and $8.8\%$, compared with
the SAE's $4.9\%$. This indicates that neither
nonlinearity, non-negativity, nor loading-side sparsity alone achieves the
SAE's code-level within-basket decorrelation.
The naive contrarian delivers the highest pre-cost rank-scale Sharpe
($+6.20$) at substantially higher volatility; it has higher Sharpe than all
factor models and serves as the no-incremental-Sharpe
benchmark in \S\ref{subsec:scope}.
ICA is omitted from Table~\ref{tab:portfolio} because its contrarian Sharpe is
$-3.07$ at $K{\ge}64$, reflecting a momentum-type signal that reverses the
contrarian direction.

\begin{table}[t]
\centering
\caption{Exploratory contrarian portfolios at matched $K{=}64$ on the crypto
static test split. Rank-scale values support cross-method comparison only; all
values are pre-cost. Bold marks the best factor model; PCA uses exact SVD.}
\label{tab:portfolio}
\footnotesize
\setlength{\tabcolsep}{4pt}
\begin{tabular}{lrrr}
\toprule
Model & Return & Volatility & Sharpe \\
\midrule
Naive            & $+47.9$ & $7.7$           & $+6.20$ \\
\midrule
SAE-$K{=}64$     & $+13.9$ & $\mathbf{4.2}$  & $\mathbf{+3.28}$ \\
PCA-$K{=}64$     & $+18.1$ & $6.8$           & $+2.67$ \\
DenseAE-$K{=}64$ & $+16.3$ & $6.4$           & $+2.53$ \\
NMF-$K{=}64$     & $+14.1$ & $7.0$           & $+2.02$ \\
\bottomrule
\end{tabular}
\end{table}

We obtain exploratory out-of-sample evidence from a five-fold walk-forward in
which the SAE is retrained from scratch and PCA-$K{=}64$ is recomputed on each
fold's training window.
The volatility advantage is consistent: SAE realised volatility ranges
$4.1$--$5.7\%$ against PCA-$K{=}64$'s $6.1$--$9.1\%$, a $32$--$37\%$
reduction in every fold.
The SAE--PCA Sharpe gap is positive on average but not statistically significant
at $n{=}5$: mean $+0.82 \pm 0.81$, $4/5$ folds, paired $t(4){=}2.25$,
$p{=}0.088$ ($5/5$ folds and $p{=}0.048$ when missing assets are masked
in place of zero-filled).
The solver matters for this comparison: a randomised SVD moves the PCA
Sharpe by about $\pm 0.2$ per fold across draws, enough to change the fold
count, so we fit PCA by exact SVD.
Against DenseAE ($+1.16 \pm 0.66$, $p{=}0.017$) the SAE leads at identical
architecture and capacity, which supports the $\ell_1$ penalty's role in the
reduction.

Against the five other structured baselines the SAE Sharpe gap is positive
in $5/5$ folds (DenseAE $+1.16$, NMF $+1.29$, ICA $+1.86$, sparse PCA
$+0.85$, dictionary learning $+1.48$; raw paired
$p{=}0.017$, $0.041$, $0.024$, $0.024$, $0.009$).
Benjamini--Hochberg false-discovery-rate (BH-FDR) control
($q{=}0.05$)~\cite{harvey2016cross} retains these five and drops PCA;
strict Bonferroni family-wise correction rejects none (smallest adjusted
$p{=}0.055$), reflecting the limited leverage of $n{=}5$ folds.
A structurally different ten-fold expanding design uses a $994$-day initial
training window, $99$-day test folds, and one rotated seed per fold. We treat
this as a separate robustness protocol, not as a higher-powered extension of
the five-fold analysis. Under strict Bonferroni correction, the SAE differs
significantly from DenseAE and dictionary learning (both paired
$t(9){=}3.67$, raw $p{=}0.005 < 0.05/6$). BH-FDR retains five of the six
comparisons, including PCA ($+1.32 \pm 1.55$, $7/10$ folds, $p{=}0.025$),
and drops ICA ($p{=}0.069$).
Fold-level volatilities appear in Table~\ref{tab:robustness}.

\begin{table}[t]
\centering
\caption{Exploratory $K{=}64$ ten-fold robustness analysis ($994$-day initial training
window, $99$-day test folds, rotating seeds). Volatility is annualised on the
rank scale; $\Delta\rho_{\mathrm{lf}}$ is the leak-free trailing $60$-day
long-basket correlation gap (SAE $-$ PCA).}
\label{tab:robustness}
\footnotesize
\setlength{\tabcolsep}{5pt}
\begin{tabular}{crrrr}
\toprule
Fold & SAE Vol. & PCA Vol. & DenseAE Vol. & $\Delta\rho_{\mathrm{lf}}$ \\
\midrule
0 & $6.6$ & $9.7$ & $10.2$ & $-0.048$ \\
1 & $5.5$ & $8.5$ & $8.4$  & $-0.032$ \\
2 & $5.2$ & $7.3$ & $7.3$  & $-0.025$ \\
3 & $5.9$ & $9.7$ & $9.7$  & $-0.039$ \\
4 & $4.5$ & $7.4$ & $7.1$  & $-0.030$ \\
5 & $4.2$ & $7.1$ & $7.2$  & $-0.030$ \\
6 & $4.0$ & $6.4$ & $6.1$  & $-0.032$ \\
7 & $4.2$ & $6.3$ & $6.2$  & $-0.024$ \\
8 & $5.2$ & $8.5$ & $8.2$  & $-0.035$ \\
9 & $5.9$ & $9.1$ & $9.0$  & $-0.036$ \\
\midrule
mean & $5.1$ & $8.0$ & $7.9$ & $-0.033$ \\
\bottomrule
\end{tabular}
\end{table}

\paragraph{Raw-return robustness.}
Signals are formed on clipped inverse-normal rank scores
(Eq.~\ref{eq:rankit}), so a
natural concern is whether the volatility reduction is an artefact of the
rank transform or persists when portfolio returns are measured in money units.
As an exploratory robustness check, we recompute the matched-$K{=}64$
contrarian long-short volatility on the
\emph{raw} winsorised log-return panel: signals are still formed on rank
scores, but portfolio profit-and-loss, realised volatility, and the
per-day eligibility mask are computed on raw returns. Basket membership uses
only signals and return availability observed on day $t$; a held asset with a
missing day-$t{+}1$ return is treated as flat in the headline calculation.
The reduction persists in money units: SAE raw annualised long-short
volatility is $20.0\%$ against PCA-$K{=}64$'s $27.3\%$, a mean reduction of
$27\%$ across the five folds (per-fold $22$--$32\%$).
Across the five \emph{disjoint} $159$-day test windows, the reduction is
positive in every fold. Because the expanding training windows create some
dependence across folds, we emphasise only that the reduction has the same
direction in all five folds.
The raw-return result is consistent with the within-basket decorrelation
observed on the rank scale and shows that the reduction is not solely a
mechanical consequence of the bounded rank scale. SAE volatility in money
units is also $17$--$34\%$ below that of the naive contrarian in all five folds.

\paragraph{Money-scale Sharpe and missing-return sensitivity.}
Under the same day-$t$ eligibility rule, mean pre-cost Sharpes are $+4.74$
for naive, $+2.26$ for SAE, and $+1.19$ for PCA. The SAE$-$naive gap is
$-2.48$ ($0/5$ folds, $t(4){=}{-}4.81$, $p{=}0.009$), so the naive ordering
of \S\ref{subsec:scope} holds in money units. The SAE$-$PCA gap is $+1.07$
($5/5$ folds, $p{=}0.084$).
To bound the flat-return convention, we assign more adverse outcomes to held
assets that disappear from the data on day $t{+}1$.
At most $67$ of ${\sim}136{,}000$ basket slots per method are affected.
If every vanished asset collapses by $50\%$, none of the three Sharpes moves
by more than $0.08$.
In a worst case, assets held long lose $50\%$ and assets held short double in
price. The three Sharpes then fall to $3.73$, $1.34$, and $0.43$. The ordering
and both gaps remain unchanged.

\begin{figure}[t]
  \centering
  \includegraphics[width=\linewidth]{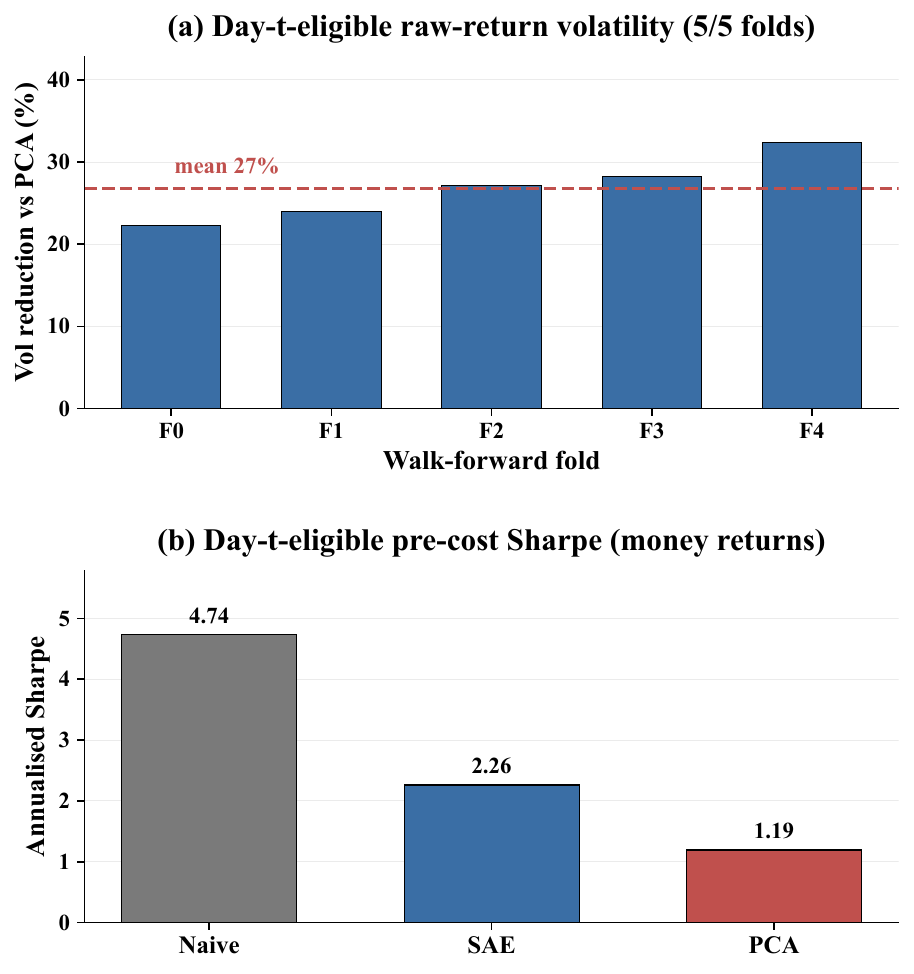}
  \caption{Exploratory day-$t$-eligible raw-return results at matched
    $K{=}64$. \textbf{(a)}~SAE volatility reduction relative to PCA is positive
    in all five folds. \textbf{(b)}~Pre-cost annualised Sharpe; the naive
    contrarian leads both reconstruction models.}
  \label{fig:pareto}
  \Description{Panel a is a five-bar chart showing a positive SAE volatility
    reduction relative to PCA in every walk-forward fold. Panel b compares
    pre-cost raw-return Sharpe ratios for the naive, SAE, and PCA portfolios.}
\end{figure}

\paragraph{Economic scope.}
Because the daily-close design is an idealised signal evaluation rather than
an executable strategy, we do not estimate monetary transaction costs or
trading capacity.
Figure~\ref{fig:pareto}b instead reports the pre-cost Sharpe ordering on raw
returns: the naive contrarian leads, followed by the SAE and PCA.

\subsection{Exploratory \texorpdfstring{$K{=}64$}{K=64} mechanism:
within-basket diversification}
\label{subsec:mechanism}

\begin{figure}[t]
  \centering
  \includegraphics[width=\linewidth]{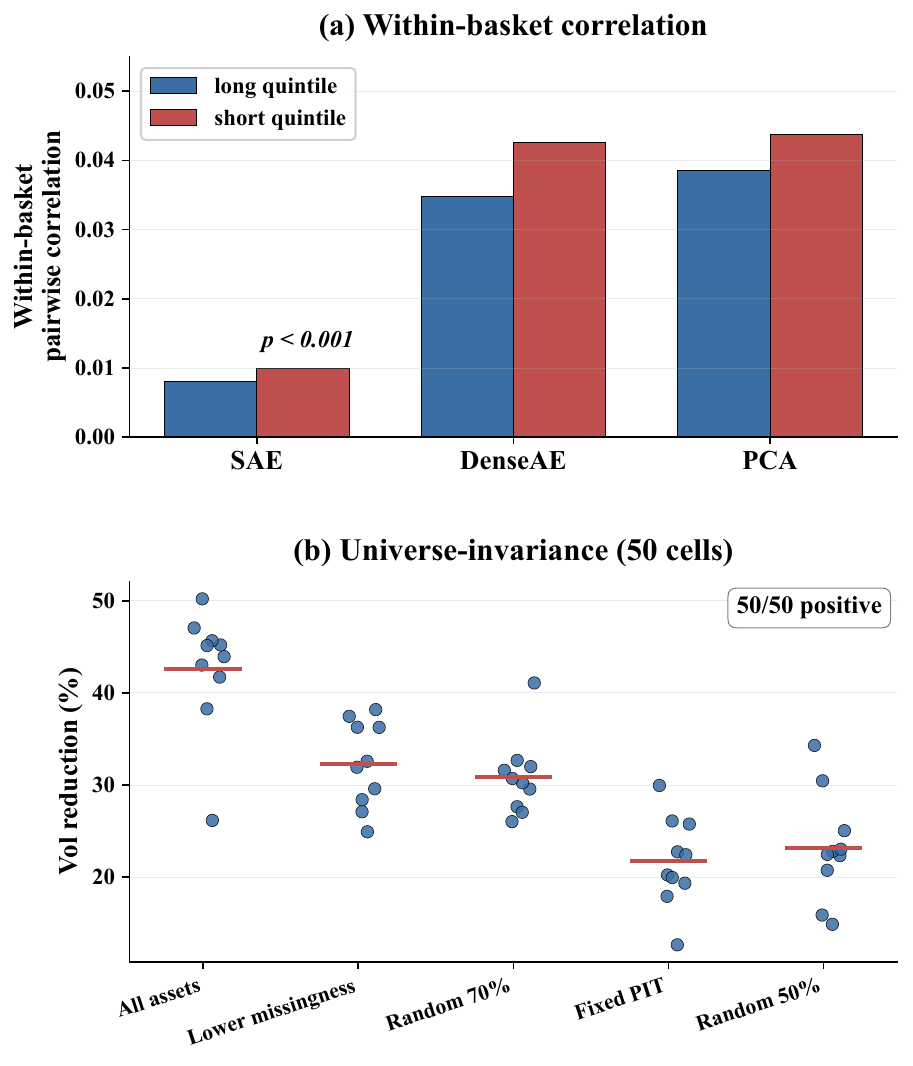}
  \caption{Exploratory $K{=}64$ diagnostics. \textbf{(a)}~Mean within-basket
    pairwise correlations. \textbf{(b)}~SAE--PCA volatility reduction across
    five universe definitions and ten folds; cells share underlying days.}
  \label{fig:mechanism}
  \Description{Panel a compares long- and short-basket correlations for SAE,
    DenseAE, and PCA, with SAE lowest on both sides. Panel b shows positive
    SAE volatility reduction in every fold under five universe definitions.}
\end{figure}

At matched $K{=}64$, the mean pairwise return correlation within SAE long
baskets is $0.008$, compared with $0.039$ for PCA. On the short side, the
corresponding values are $0.010$ and $0.044$. The PCA correlations are
$4.4$--$4.8$ times as large as the SAE correlations ($30$-day stationary
block-bootstrap $p{<}0.001$, $n{=}368$ rebalance days).
These are full-window descriptive statistics; a leak-free out-of-sample check
follows below.
Both factor models produce within-basket correlations above the
universe-average reference ($\bar\rho_{\text{univ}} {=} 1.28\times10^{-4}$,
$N{=}433$), but the increase is much smaller for the SAE. This pattern is
consistent with an $\ell_1$ penalty that selects assets from less-correlated
return directions.
The associated covariance decomposition is informative.
Writing the long-short variance as
\[
  \mathrm{Var}(L{-}S) = \mathrm{Var}(L) + \mathrm{Var}(S) - 2\,\mathrm{Cov}(L,S),
\]
and splitting each leg into a diagonal (idiosyncratic) and an off-diagonal
(within-basket) part yields four terms.
On the static split, the SAE--PCA portfolio-variance difference is $-0.029$
rank units, a $32\%$ reduction ($p{<}0.001$). Within-basket covariance
accounts for $99.5\%$ of this difference: $46.7\%$ comes from the long leg and
$52.8\%$ from the short leg. Cross-basket hedging contributes another
$36.2\%$, whereas the idiosyncratic component offsets $35.6\%$ of the
reduction. The idiosyncratic component has the opposite sign because the SAE
holds higher-variance assets, contrary to a low-volatility-selection
explanation. It roughly cancels the gain from cross-basket hedging. Thus,
lower within-basket covariance is the dominant component of the static
SAE--PCA variance difference. All four components are significant under the
$30$-day stationary block bootstrap ($p{<}0.001$, $n{=}368$).
This covariance attribution uses a single seed and a trailing $30$-day window,
so its scale differs from that of ex-post realised volatility. It explains
most of the static variance difference associated with the $4.2\%$ versus
$6.8\%$ volatility result in Table~\ref{tab:portfolio}. The volatility
reduction remains positive in all $50$ combinations of universe definition and
fold, as well as all $40$ random sub-panel draws
(Fig.~\ref{fig:mechanism}b). Across the five universe definitions and ten
folds, reductions range from approximately $13\%$ to $50\%$.

Lower portfolio volatility could arise from two channels: selecting
low-idiosyncratic-volatility assets, or decorrelating a basket of
higher-volatility assets.
An asset-level analysis does not support the former explanation. SAE basket
members have idiosyncratic volatility $2\%$ above the universe mean, and
the SAE$-$PCA basket idiosyncratic-volatility difference is $+2.7$pp
($30$-day stationary block-bootstrap $p{<}0.001$, $n{=}368$ rebalance days),
the opposite of what the low-volatility-selection explanation predicts.
The SAE baskets contain higher-variance assets but exhibit lower pairwise
return correlations.

We next estimate the correlation gap out of sample using only a trailing
$60$-day basket-correlation window.
Sharpe inference (\S\ref{subsec:portfolio}) uses the primary five-fold
walk-forward, while the mechanism's leak-free check uses the finer ten-fold
split for more precise per-day correlation estimates. The ten-fold design yields
long-side $\Delta {=} -0.033$, negative in $10/10$ folds (per-fold values in
Table~\ref{tab:robustness}). Because the trailing windows overlap heavily, we report
the fold-level sign count rather than a pooled daily $t$-test. The five-fold
variant agrees in direction (long-side $\Delta {=} -0.007$, negative in all
five folds).

\paragraph{$\lambda$-ablation: decorrelation without underfitting.}
A seven-point $\lambda$ sweep at an otherwise fixed specification
($K{=}64$, identical optimiser settings, static test split) separates
the bottleneck mechanism from sparsity-induced underfitting.
For $\lambda \in [10^{-4}, 5{\times}10^{-2}]$, reconstruction mean squared
error (MSE) is
nearly constant ($1.043$--$1.047$, spread $<0.5\%$), while trailing $60$-day
long-basket correlation is only about one-fifth to one-third of the
$\lambda{=}0$ DenseAE value ($0.0066$--$0.0108$ vs $0.0328$). Long-short
volatility is $21$--$30\%$ lower. Only at $\lambda{=}10^{-1}$ does
reconstruction degrade (MSE $1.18$) with no decorrelation gain.
Decorrelation therefore appears once $\lambda>0$ in the ReLU-plus-$\ell_1$
code, with little reconstruction cost. This pattern argues against
underfitting. Within the nearly constant-MSE range, correlation increases
slightly with $\lambda$, so we describe a threshold effect rather than a
monotone dose-response.

\subsection{Scope: overlay control and no incremental Sharpe}
\label{subsec:scope}

At the exploratory $K{=}64$ setting, we ask whether the volatility
advantage requires a sparse autoencoder or whether a simple diversification
overlay reproduces it.
We address this with an additional control: a
Ledoit--Wolf~\cite{ledoit2004well} minimum-variance overlay applied to
the naive contrarian's quintile portfolios.
On the five-fold walk-forward, the strengthened naive baseline attains
realised annualised volatility $5.60\%$ versus the SAE's $4.85\%$,
reducing the SAE--naive volatility gap by approximately $73\%$. A smaller
descriptive difference remains in four of five folds; the fifth shows no clear
difference (paired $t(4){=}{+}2.65$, one-sided $p{=}0.029$, two-sided
$p{=}0.057$).
The margin is crypto-specific: on equities, the overlaid naive portfolio is
less volatile than the SAE portfolio in all five folds ($6.2\%$ vs $8.6\%$).
Most of the volatility advantage is therefore reproducible by standard
covariance shrinkage on the naive signal. The overlay does not remove the
difference completely, but the remaining evidence is not conclusive under a
two-sided test. We therefore do not interpret the remainder as a separate
mechanism.

Beyond volatility, the SAE gains no incremental Sharpe over stronger naive
baselines.
The naive equal-weighted contrarian achieves higher pre-cost rank-scale Sharpe
than the SAE (SAE--naive gap $-2.91$, stationary block-bootstrap $95\%$
confidence interval
$[-4.40,-1.34]$, $p{=}0.002$)~\cite{politis1994stationary} and in money
units ($-2.48$ walk-forward, $p{=}0.009$, \S\ref{subsec:portfolio}), and a
lagged-return MLP (LR-MLP) in the spirit of
\citet{gu2020empirical} (per-asset MLP on a $20$-day trailing return
window, trained per fold) widens the gap further: the SAE's residual
Sharpe orthogonal to the joint (naive, LR-MLP) span is $-5.98$
(paired $t{=}-7.94$, $p{<}0.001$).
These results characterise the SAE primarily as a volatility regulariser and
are consistent with the within-basket decorrelation evidence; its risk-adjusted
return remains below that of the naive contrarian.
This scope statement carries into the cross-asset results
(\S\ref{subsec:cross-asset}) and the conclusion.

\begin{table}[t]
\centering
\caption{Summary of empirical findings. Strict confirmation applies only to the $K{=}128$
volatility result; the $K{=}64$ evidence is exploratory.}
\label{tab:claims}
\footnotesize
\setlength{\tabcolsep}{4pt}
\begin{tabular}{>{\raggedright\arraybackslash}p{0.30\linewidth}
                >{\raggedright\arraybackslash}p{0.34\linewidth}
                >{\raggedright\arraybackslash}p{0.22\linewidth}}
\toprule
Claim & Evidence & Status \\
\midrule
Volatility reduction
  & Strict $K{=}128$: $20$--$26\%$, $5/5$ folds; exploratory $K{=}64$:
    $32$--$37\%$, $5/5$
  & supported \\[2pt]
Within-basket decorrelation
  & Exploratory $K{=}64$: leak-free $\Delta {=} {-}0.033$, $10/10$ folds
  & supported \\[2pt]
Low-volatility selection
  & $K{=}64$ SAE--PCA idiosyncratic volatility $+2.7$pp, $p{<}0.001$
  & not supported \\[2pt]
Sharpe against structured baselines
  & Exploratory $K{=}64$ suggestive; strict $K{=}128$: $+0.33$, $p{=}0.14$
  & suggestive \\[2pt]
Incremental Sharpe against naive
  & Rank-scale gap $-2.91$; day-$t$-eligible raw-return gap $-2.48$
  & not supported \\[2pt]
Seed-reproducible identities
  & Matched cosine $0.13$ compared with random $0.12$; best match $0.23$
  & not supported \\[2pt]
Cross-asset Sharpe transfer
  & Equity SAE--PCA gap $-0.20$, $p{=}0.160$; SAE also trails naive
  & crypto-specific \\
\bottomrule
\end{tabular}
\end{table}

\subsection{Cross-asset transfer and factor correlations}
\label{subsec:cross-asset}

As a secondary exploratory check, we apply the method to a $200$-stock
S\&P~500 panel from 2015--2024. The SAE again has the lowest realised
long-short volatility of any factor-model method: $6.6\%$, which is $21\%$
below matched-$K$ PCA's $8.4\%$ on the static test split. This result extends
the volatility-reduction pattern to a structurally different asset class,
although the equity panel remains exposed to survivorship bias
(\S\ref{subsec:data}).
The transfer holds fold-by-fold under the equity five-fold walk-forward. SAE
realised volatility is $9.9\%$, $9.6\%$, $8.4\%$, $7.6\%$, and $7.5\%$; the
corresponding PCA values are $12.2\%$, $11.7\%$, $9.5\%$, $8.8\%$, and $8.4\%$.
The SAE is lower in all five folds, with reductions of $10$--$19\%$.
The SAE factors have low marginal pairwise correlation with the canonical
Fama-French five-factor model plus momentum~\cite{fama2015five}. On equities,
the largest absolute correlation is $0.147$, so no single classical factor
explains them. This is a marginal check, not a joint spanning test.
However, the crypto Sharpe edge fails to transfer: the equity
five-fold SAE$-$PCA Sharpe difference is not statistically significant
($-0.20$, $p{=}0.160$), and the naive
contrarian again leads (SAE $+0.80$ vs naive $+1.46$; SAE$-$naive gap
$-0.66$, $p{=}0.072$), consistent with the
volatility-regularisation reading of \S\ref{subsec:scope}.
On crypto, the SAE factors likewise show low marginal correlation with
FF5+MOM (max $|\rho|{=}0.157$). This is only a pairwise check, not a spanning
test against crypto-native factor portfolios.

\subsection{Interpretability: no reproducible dictionary}
\label{subsec:interp}

A core motivation for the SAE architecture is the monosemanticity reported
in NLP~\cite{bricken2023monosemanticity}. At the exploratory $K{=}64$
setting, we ask whether individual factor identities are stable:
similar directions should reappear when the model is retrained with another
seed. This is a reproducibility diagnostic, not a complete test of
semantic coherence, feature splitting, or subspace stability. In each
walk-forward fold we align the decoder columns of the five seed models
pairwise by Hungarian matching on absolute cosine similarity.
The mean matched cosine is $0.13$ in every fold.
Pairs of random dictionaries of the same shape score $0.117 \pm 0.002$. The
observed alignment is above this random reference but remains small: all $50$
seed pairs lie above its maximum over $500$ draws, yet the best match among
$3{,}200$ factor pairs is only $0.23$.
Each seed therefore learns a different dictionary under this matching
criterion, while the portfolio effect is seed-stable (all $25$
fold$\times$seed cells). We find no seed-reproducible per-factor identity;
the test does not rule out meaningful features within one run or a stable
higher-dimensional subspace.

As a descriptive exercise, we use the Llama-3.1-8B large language
model~\cite{grattafiori2024llama} to assign labels to the $64$ factors of one fitted
SAE from their top-loading assets. The automated labels vary little across
factor representations. Applied to SAE, PCA, DenseAE, and random-projection
factors in the final walk-forward fold, the same procedure reports almost the
same mean confidence for all four ($0.82$, $0.81$, $0.82$, $0.81$) and the same
theme mix: momentum, volatility and liquidity account for $88$--$97\%$ of
the labels in every representation.
After aligning factors across seeds, SAE theme agreement ($0.28$--$0.44$)
exceeds the $95$th percentile of a permutation null ($0.42$) in one of
three seed pairs.
We make no monosemanticity claim; a stronger claim would require both
cross-seed stability and human-validated concept naming.

\begin{figure}[t]
  \centering
  \includegraphics[width=\linewidth]{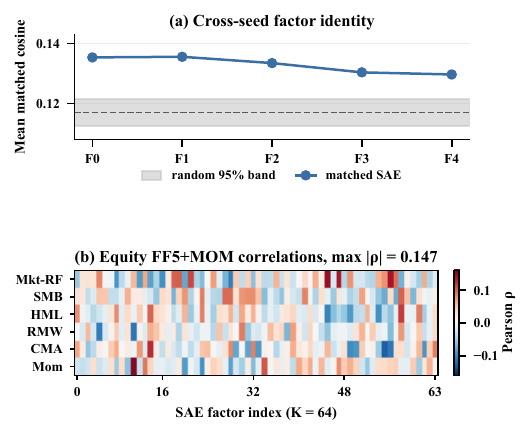}
  \caption{Factor stability and correlations. \textbf{(a)}~Mean Hungarian-matched
    decoder cosine by fold; the shaded band is the random-dictionary $95\%$
    reference interval.
    \textbf{(b)}~Descriptive equity-sample correlations with FF5+MOM
    ($\max|\rho|{=}0.147$), short of a joint spanning test.}
  \label{fig:factorzoo}
  \Description{Panel a compares the mean matched decoder cosine in each fold
    with a narrow random-dictionary reference band; the observed values are
    only slightly above random. Panel b is a heat map of small positive and negative
    correlations between 64 equity-sample SAE factors and six FF5 plus momentum
    series.}
\end{figure}

\section{Discussion and Conclusion}
\label{sec:discussion}

The $\ell_1$ bottleneck behaves as an implicit diversification regulariser for
cross-sectional return panels.
The exploratory mechanism evidence points to within-basket decorrelation
(\S\ref{subsec:mechanism}):
pairwise return correlations inside PCA baskets are $4.4$--$4.8$ times as
large as those inside SAE baskets at matched $K{=}64$.
Low-volatility asset selection is not supported by the asset-level evidence; after a
Ledoit--Wolf overlay reduces the naive gap by approximately $73\%$
(\S\ref{subsec:scope}), a smaller descriptive difference remains in four of
five crypto folds. This difference is significant under a one-sided test
($p{=}0.029$) but not under a two-sided test ($p{=}0.057$).

The naive contrarian achieves the highest pre-cost Sharpe (SAE-versus-naive gap
$-2.91$ on the rank scale and $-2.48$ in money units), and a lagged-return
MLP widens the gap further. The evidence therefore supports a specific
interpretation of the $\ell_1$ bottleneck's effect on portfolio structure:
the strict $K{=}128$ experiment confirms lower volatility, while the
exploratory $K{=}64$ analyses are consistent with the reduction arising
largely from within-basket covariance. The method provides no incremental
risk-adjusted return, seed-reproducible dictionary, or cross-asset Sharpe
improvement (Table~\ref{tab:claims}).
The exploratory matched-$K$ Sharpe edge is positive in $5/5$ crypto folds against
five structured baselines and in $4/5$ against exact-SVD PCA
($p{=}0.088$); it is specific to the setting and to crypto: on
S\&P~500 equities, the SAE trails PCA. The volatility reduction and low
correlations with classical factors do transfer to equities
(\S\ref{subsec:cross-asset}).

Four directions remain open: conditional encoders with firm
characteristics~\cite{gu2021autoencoder}; broader equity panels with
intraday execution data; the use of sparse codes to identify market regimes
and as inputs to downstream models; and a theory linking the $\ell_1$
bottleneck to basket-level decorrelation.

\balance
\bibliographystyle{ACM-Reference-Format}
\bibliography{references}

\end{document}